\documentclass{svjour3}                     % onecolumn (standard format)
\usepackage[utf8]{inputenc} 
\smartqed  % flush right qed marks, e.g. at end of proof

\usepackage{graphicx}% Include figure files
\usepackage{dcolumn}% Align table columns on decimal point
\usepackage{bm}% bold math
\usepackage{amssymb,amsbsy,exscale,amsmath,amsfonts,amscd}
\usepackage{graphicx}
\usepackage[letterpaper]{geometry}
\usepackage[usenames,dvipsnames]{color}
\usepackage{epstopdf}
\usepackage{stmaryrd}
\usepackage{mathrsfs}
\usepackage{upgreek}
\usepackage[compress,numbers]{natbib}
\usepackage[font=footnotesize]{caption}
\usepackage{array}
\usepackage{parskip}
\usepackage{mathtools}
\usepackage{textcomp}
\usepackage[load-configurations = abbreviations]{siunitx}
\newcommand{\abs}[1]{\left\lvert{#1}\right\rvert}

\begin{document}
\title{Automatic event detection in football using tracking data}

\author{Ferran Vidal-Codina    \and 
        Nicolas Evans     \and
        Bahaeddine El Fakir \and
        Johsan Billingham
}

%\authorrunning{Short form of author list} % if too long for running head

\institute{F.~Vidal-Codina \at
              MIT Sports Lab, Massachusetts Institute of Technology, Cambridge, MA 02139, USA \\
              \email{fvidal@mit.com}           %  \\
           \and
          N.~Evans, B.~El~Fakir, J.~Billingham  \at
              Football Research and Standards, FIFA, CH-8037 Z\"{u}rich, Switzerland
}

%\author{}

\date{Received: date / Accepted: date}

\maketitle

\begin{abstract}
One of the main shortcomings of event data in football, which has been extensively used for analytics in the recent years, is that it still requires manual collection, thus limiting its availability to a reduced number of tournaments. In this work, we propose a deterministic decision tree-based algorithm to automatically extract football events using tracking data, which consists of two steps: (1) a possession step that evaluates which player was in possession of the ball at each frame in the tracking data, as well as the distinct player configurations during the time intervals where the ball is not in play to inform set piece detection; (2) an event detection step that combines the changes in ball possession computed in the first step with the laws of football to determine in-game events and set pieces. The automatically generated events are benchmarked against manually annotated events and we show that in most event categories the proposed methodology achieves $+90\%$ detection rate across different tournaments and tracking data providers. Finally, we demonstrate how the contextual information offered by tracking data can be leveraged to increase the granularity of auto-detected events, and exhibit how the proposed framework may be used to conduct a myriad of data analyses in football.
% \begin{keyword}
% football analytics \sep tracking data \sep event data \sep ball possession \sep auto-eventing \sep data democratization
% \end{keyword}
\end{abstract}

\section{\label{sec:intro}Introduction}

Teams, media, experts and fans have always analyzed football and try to best explain what is happening on the pitch. Until recently, broadcast footage was the only true source of information, leading to qualitative analysis based on observation. As the first analysis software appeared and information technology matured, thus enabling real-time data transmission, it became possible to bookmark events of interest in the game for several purposes --coaching, highlight editing or third party applications. From this, a standardized dataset known as \textit{event data} emerged. Event data has traditionally been a file containing all manually collected events that occurred in a given football match. These datasets are nowadays used by most stakeholders in football, some of which are starting to find limitations due to the nature of the dataset. 

Furthermore, in the recent years there has been a rise in demand for \textit{tracking data}, namely technologies providing center-of-mass coordinates for all players and the ball several instances per second, obtained through Electronic Performance \& Tracking Systems (EPTS) \cite{fifaepts}. These ever-improving systems have motivated the appearance of new opportunities to quantify many of the observations that had previously only been qualitative --one of those being the determination of events. With a view of making technology more globally available, FIFA started a research stream to analyze whether the events could be identified using tracking data (and potentially computer vision), thus eliminating the need for manual coding. With a vision of extracting tracking data from broadcast footage in the foreseeable future, the overarching objective of this research is to be able to provide video, tracking and event data from a single camera going forward, thus contributing to the development of the game while improving the consistency and repeatability of the event data collection.

Event data for major football leagues and tournaments started to be collected by Opta Sports (now Statsperform) \cite{statsperform} at the end of the 20th century, and its widespread adoption has propelled the development of advanced football statistics for analytics, broadcast and sports-betting \cite{wang2015discerning,marchiori2020secrets,van2021leaving,montoliu2015team,szczepanski2016beyond,gyarmati2014searching,bekkers2019flow,gonccalves2017exploring,lucey2012characterizing,brooks2016using,decroos2019actions,tuyls2021game,decroos2020player}. There are now various commercial providers that manually collect event data for different leagues \cite{statsperform,stswebsite,wyscout,statsbomb}, event data from past seasons is widely available. Nonetheless, collection of event data presents some challenges: (1) since it serves different purposes, there is a lack in consistency in both terminology and event definitions, as well as granularity and accuracy of the time annotation; (2) the nature of the task is subjective, hence there may be substantial tagging differences (10-15\%) among analysts; (3) event data needs to be post-processed for quality control, thus the final dataset for some providers may not typically available until several hours after the match; (4) only information for the player executing the event is available, thus there is no information on the broader game context (\textit{e.g.} location of other players at time of event) --although some providers \cite{statsbomb} have recently started including this information; and (5) manual data collection is resource-intensive, and thus cannot be readily extended to the majority of football tournaments.

To address the latter, there have been many efforts in the field of computer vision to automatically detect events using broadcast video \cite{ekin2003automatic,d2010review,assfalg2003semantic,tavassolipour2013event,kapela2014real,ballan2009action}. More recently, convolutional and recurrent neural networks have been employed for this task \cite{giancola2018soccernet,baccouche2010action,jiang2016automatic,tsagkatakis2017goal,deliege2021soccernet,rongved2020real,cioppa2018bottom}, which has been enabled by the extensive availability of manually tagged datasets and the recent advances in action recognition. However, automatic event detection with video has thus far focused on a subset of football events, namely goals, shots, cards, substitutions and some set pieces, devoted mainly to highlight generation. Therefore, the automatically generated event logs are sparse and lack game context, thus limiting its applicability for advanced football analytics and granular game description. 

The absence of game context on event data has been partially addressed in the recent years with the advent of tracking data. In particular, tracking data collected with optical EPTS is the most common due to its accuracy (which has benefited from the recent advances in deep learning and computer vision methodologies) and minimal invasiveness \cite{fifaepts}. There are a myriad of commercial providers that collect football tracking data using optical systems for clubs, leagues and federations \cite{metricawebsite,tracabwebsite,t160website,kogniawebsite,sswebsite,heiwebsite}. The main drawback of optical EPTS is the need for a camera installation in every stadium, albeit there are promising research and commercial avenues \cite{metricawebsite,sportlogiq,footovision,skillcorner} to extract tracking data from broadcast or tactical footage that mitigate the costs. Another notable drawback of optical EPTS is that the data quality depends on the stadium, namely the height at which the cameras are installed.

Tracking data provides a richer context than event data, since information on all players, their trajectories and velocities is readily available, which enables the evaluation of off-ball players and team dynamics. Consequently, storing and computing with tracking data is more resource-intensive than with event data --a football match at $\SI{25}{\Hz}$ contains roughly 3 million tracking data points, compared to 3K events. In the recent years, tracking data has been extensively used to perform football analytics, and its proliferation has given rise to several advanced metrics, for instance the quantification of team tactics, pitch control or expected possession value to name a few \cite{lucey2014quality,le2017data,bialkowski2014identifying,gudmundsson2014football,spearman2018beyond,shaw2020routine,power2017not,link2016real,cakmak2018computational,fernandez2019decomposing,dick2019learning,fernandez2018wide,stockl2022making}. 

To the best of our knowledge, this article represents the first attempt to use tracking data as a means to automatically generate event data. There are many advantages to this approach: (1) it generates event data that would otherwise need to be manually annotated; (2) tracking data has been automatically extracted from video, thus containing highly-curated information on players and ball; (3) events generated from tracking data are not only synced with tracking data, but also highly specific, since information on the other players and the match context is available; and (4) combining automatic event generation with tracking data from broadcast/tactical footage will further the democratization of tracking and event data. 

Here, we propose to extract possession information and football event data from 2D player and ball tracking data. To that end, we have developed a deterministic decision tree-based algorithm that evaluates the changes in distance between the ball and the players to generate in-game events, as well as the spatial location of the players during dead ball intervals to detect set pieces, hence there is \textbf{no learning involved}. The output consists of a chronological log of possession information and discrete events. This article is organized as follows. In Section \ref{sec:Methods}, we describe in detail the proposed algorithm and the different datasets used. In Section \ref{sec:results}, we benchmark the automatically generated events against manually annotated events and showcase how the auto-eventing algorithm can be used for football analytics. In Section \ref{sec:discussion}, we discuss the benchmarking results, limitations of the algorithm and the data and perspectives for future research.

\section{Materials and Methods}\label{sec:Methods}
\subsection{Data resources}\label{sec:data}
We have used tracking data from three different tracking data providers: Track160 \cite{t160website} (hereafter referred to as provider A), Tracab \cite{tracabwebsite}  (hereafter referred to as provider B) and Hawk-Eye \cite{heiwebsite}  (hereafter referred to as provider C) across three tournaments as follows: for Track160, six games in the FIFA Club World Cup 2019 (FCWC19) provided by FIFA and three games in the 2019-2020 Bundesliga season provided by Track160; for Tracab, seven games (three of them processed with version 5.0 and the remaining four withprovider A version 4.0) in the FIFA Club World Cup 2020 (FCWC20) provided by FIFA  and twelve games (version 4.0) in the 2019-2020 Men's Bundesliga season provided by Deutsche Fussball Liga (DFL); and for Hawk-Eye, three games in the FCWC20 provided by FIFA --data for these three games was also collected with Tracab 4.0. In all cases, the tracking data consists of $(x,y)$ coordinates for all the players and the ball sampled at $\SI{25}{\Hz}$. Since the $z$-coordinate of the ball is not available for all datasets, we only use 2D ball information. In addition to the ball and player coordinates, tracking data contains information on the status of the game at every frame, either directly with a boolean that switches between in-play or dead ball (Tracab) or indirectly with missing ball data when the game is dead (Track160 and Hawk-Eye).

To benchmark the automatically detected events, we have used official event data collected by Sportec Solutions (STS) \cite{stswebsite} for all games, provided by FIFA for FCWC19-20 and by DFL for the Bundesliga games. These official events are indexed by game, half, minute, second and player or players that executed the event.

All data subjects were informed ahead of collection that "Optical player tracking data, including limb-tracking data, will be collected, and used for officiating, performance analysis, research, and development purposes" thus providing the basis for legitimacy of use in this research study. The authors received human research ethics approval to conduct this work from the Committee on the Use of Humans as Experimental Subjects (COUHES-MIT).

\subsection{Computational framework}\label{sec:framework}
\begin{figure}[h!]
    \centering
    \includegraphics[scale=1]{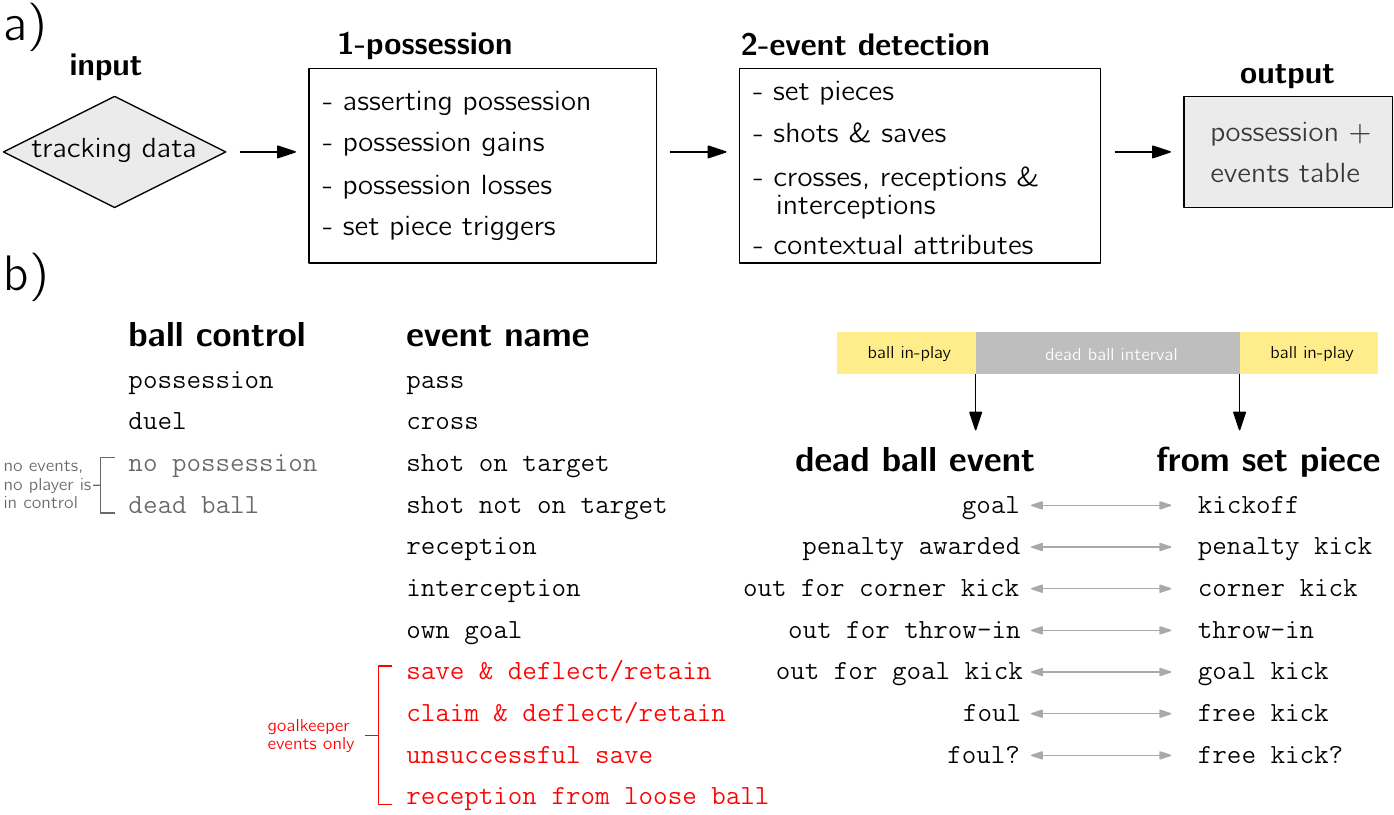}
    \caption{a) Proposed computational framework, along with information generated at each step. b) Schematic detailing all possible labels for the attributes \textbf{ball control}, \textbf{event name}, \textbf{dead ball event} and \textbf{from set piece} on the output events table.}
    \label{fig:framework}
\end{figure}

We propose a two-step algorithm to detect events in football using 2D player and ball tracking data, see Fig. \ref{fig:framework}a for a depiction of the algorithm's flowchart where all the relevant information generated at each step is detailed. The input is a tracking data table for a given game, formatted as one entry per player and frame (with ball data incorporated as column). 

The first step is determining ball possession, which is the backbone of the computational framework, as well as players' configuration during dead ball intervals. In the second step, we propose a deterministic decision tree based on the Laws of the Game \cite{ifabwebsite} that enables the extraction of in-game and set piece events from the possession information that has been established on the first step. 

The output of the algorithm is a table that for each frame in the tracking data contains automatically generated event data. In the output events table, besides information on time and players involved, we include the attributes \textbf{ball control}, \textbf{event name}, \textbf{from set piece} and \textbf{dead ball event}, see Fig. \ref{fig:framework}b. \textbf{Ball control} takes four possible values, that is: \texttt{dead ball}, \texttt{no possession}, \texttt{possession} and \texttt{duel} --the last two if at least one player is in close proximity of the ball. Ball control thus represents a continuous action, and since events occur only when ball control is either possession or duel, we may drop the rows where \textbf{ball control} is either \texttt{dead ball} or \texttt{no possession} for convenience. \textbf{Event name} refers to the in-game actions that occur on a discrete time: \texttt{pass}, \texttt{cross}, \texttt{shot on target}, \texttt{shot off target}, \texttt{reception}, \texttt{interception} and \texttt{own goal}. The goalkeepers feature additional events, namely \texttt{save} (\texttt{deflect} or \texttt{retain}) and \texttt{claim} (\texttt{deflect} or \texttt{retain}), \texttt{unsuccessful save} (the goalkeeper touches the ball but a goal is conceded) and \texttt{reception from loose ball}. The list of in-game events that we propose to detect is by no means comprehensive, but rather we focused on events that are both descriptive of the game and can be identified from tracking data using rules and without learning. Additional data streams, such as the $z$-coordinate of the ball, player limb tracking or video, may be leveraged to expand the automatically detectable events, for instance tackles, air/ground duels or dribbles.

\textbf{Dead ball event} is an attribute of the event immediately preceding a dead ball interval, namely, \texttt{out for corner kick}, \texttt{out for goal kick}, \texttt{out for throw-in}, \texttt{foul}, \texttt{penalty awarded} and \texttt{goal}, whereas \textbf{from set piece} is an attribute of the event (a pass, shot or cross) that resumes the game event after a dead ball interval, namely \texttt{corner kick}, \texttt{goal kick}, \texttt{throw-in}, \texttt{free kick}, \texttt{penalty kick} and \texttt{kickoff}. An additional pair \texttt{foul?}-\texttt{free kick?} is introduced to account for instances where the algorithm is confused due to inaccuracies in the tracking data, see Section S1 of Online Resource 1 for further details. %Finally, we show in Section \ref{sec:context} how tracking data may be leveraged to enrich the event information using contextual information, resulting in additional attributes to further characterize the auto-detected events.

% \begin{table}[h!]
%     \centering
%     \begin{tabular}
% {c||c||c|c}
% ball control   & event name & dead ball event & from set piece\\ \cline{1-4}
% possession&5 & 6&     \\[2em] \cline{1-4}
% duel& 4&   5 & 4 \\[2em] \cline{1-4}
% no possession  &4 & 5&5 \\[2em] \cline{1-4}
% dead ball   &4&{50} & \\[2em] \cline{1-4}
% bar&50&70& 
% \end{tabular}
%     \caption{Caption}
%     \label{tab:my_label}
% \end{table}

% Since pitch length and width $(L_0,L_1)$ may vary across stadiums, we set the origin of the 2-D space where the tracking data is defined to the center of the pitch, such that the corner marks are given by the 4 combinations of $(\pm L_0/2,\pm L_1/2)$. 

\begin{figure*}[h!]
    \centering
    \includegraphics[scale=1]{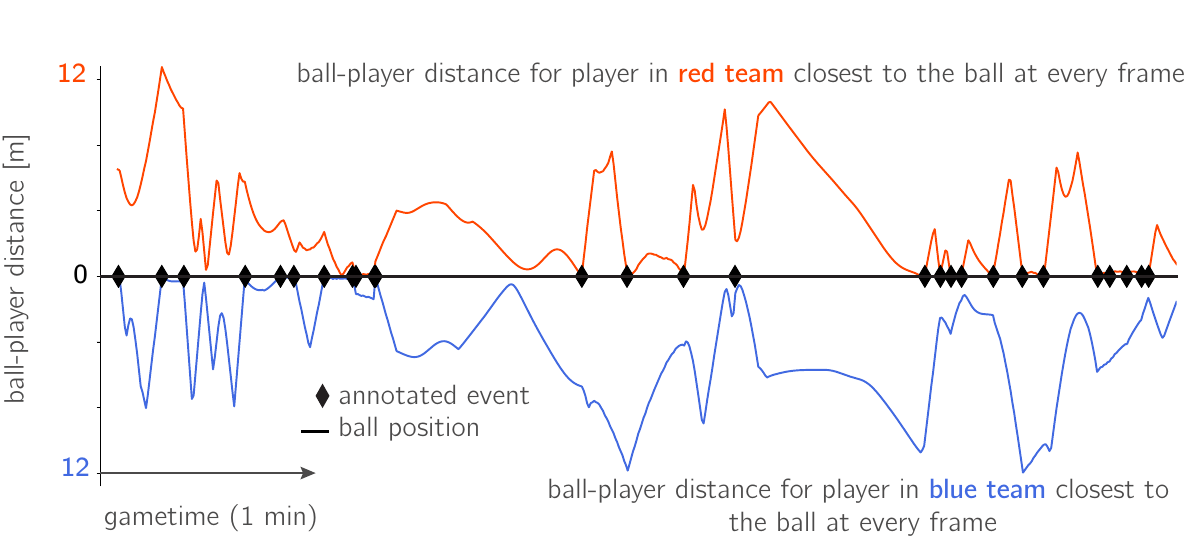}
    \caption{Distance between ball (horizontal black line) and closest player of each team (blue and red lines) for each frame within first minute of the 2019 FIFA U20 World Cup opening game, along with annotated events as black diamonds. This illustrates how in-game events occur whenever at least one player is in close proximity of the ball.}\label{fig:distance_event}
\end{figure*}

\subsection{Possession}\label{sec:possession}
\subsubsection{Asserting possession from tracking data}

Ball possession is paramount, because in-game events in football occur whenever at least one player is close to the ball, see Fig. \ref{fig:distance_event}. To establish possession, we introduce the concept of possession zone (PZ), which for simplicity we define as a circular area of radius $R_{pz}$ around every player, such that if at any given frame the ball is within a player's PZ, then that player is deemed to be in possession. Similarly, we introduce a duel zone (DZ), defined as a circular area of radius $R_{dz} \ge R_{pz}$ around the ball, such that if at least two opponents are within the DZ, then we deem there is a duel situation. 

\begin{figure*}[h!]
    \centering
    \includegraphics[scale=.8]{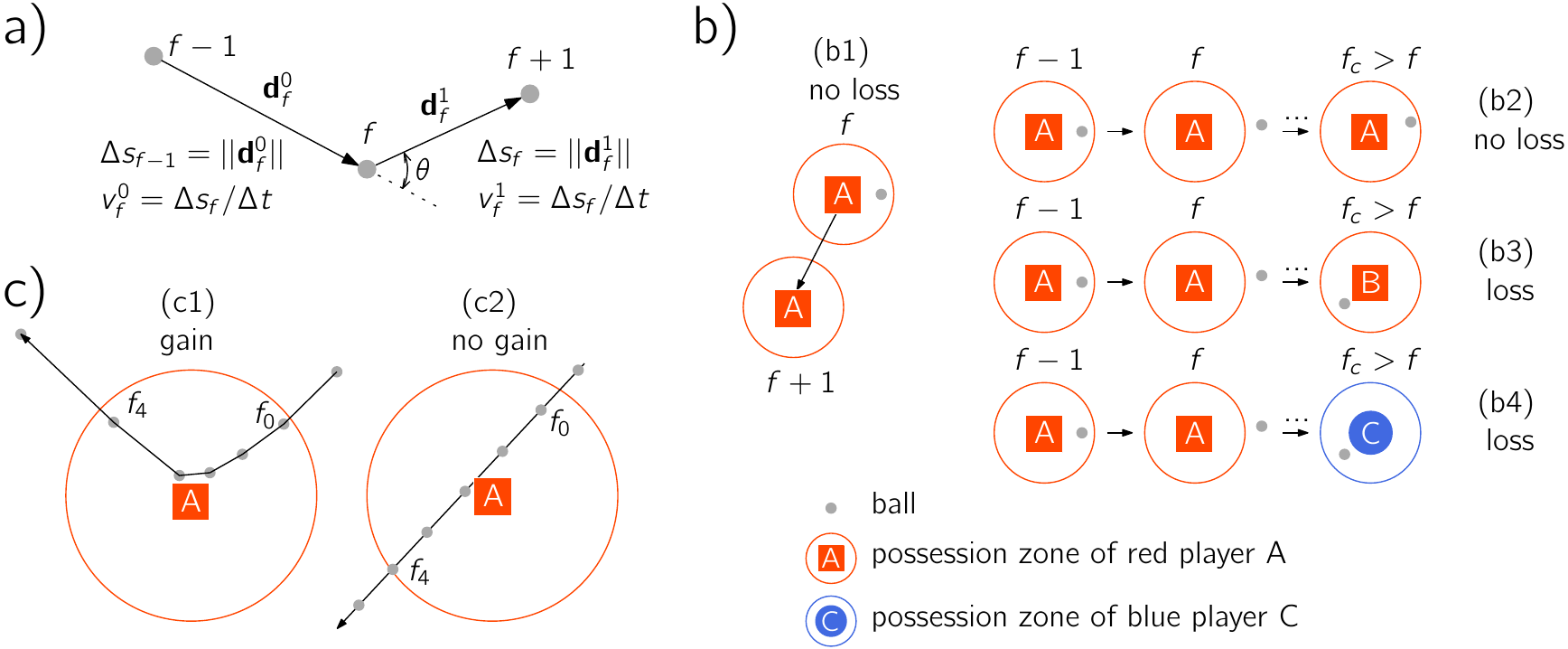}
    \caption{Schematic of ball information collection and possession losses and gains. (a) Ball information collected on each tracking data frame, including incoming/outgoing direction, speed and displacement. (b) Different potential losses, where $f$ is loss frame and $f_c > f$ is next frame where any player is in control: (b1) player moving away from static ball $\rightarrow$ no loss; (b2) player losing possession and regaining afterwards without any other player having been in control $\rightarrow$ no loss; (b3,b4) player losing possession and next player in control is either a teammate or an opponent $\rightarrow$ loss. (c) Different potential gains for a control frame interval $[f_0,f_4]$, where player is assumed static and ball position is shown in consecutive frames, moving in the direction of the arrow: (c1) ball changes trajectory and speed $\rightarrow$ gain; (c2) ball trajectory and speed remain constant $\rightarrow$ no gain.}
    \label{fig:loss_gain}
\end{figure*}

The possession algorithm reads the tracking data and applies the PZ/DZ conditions above to every frame. If both possession and duel conditions are triggered, the duel condition prevails. A frame where either possession or duel is selected is hereafter referred to as a \textit{control frame}. In addition to possession/duel information, for each control frame $f$ we store the players' distance to the ball, the ball displacement $\Delta s$ from frame $f$ to $f+1$, and the incoming ball direction vector ${\bf d}_f^0$ and speed $v_f^0$ (magnitude of velocity vector) using data from frame $f-1$ (resp. outgoing ${\bf d}_f^1$ and $v_f^1$ using data from frame $f+1$ frame), see Fig. \ref{fig:loss_gain}a. If the ball positional data has been smoothed, the speed may be computed using finite differences. Conversely, if the positional data is noisy we apply a Savitzky-Golay filter \cite{savitzky1964smoothing} to both $x-y$ ball coordinates, using a second-order polynomial and a window of seven frames around each datapoint (which for a \SI{25}{\Hz} feed corresponds to using the data from the neighboring \SI{0.25}{\s} to smooth the signal).

Once the control frames have been established, the next step is detecting changes in ball control, i.e. gains or losses, that will later be classified as in-game events by the event detection step. Gains and losses are determined upon the control frames extracted from tracking data as follows.

\subsubsection{Possession losses}
Player A loses possession at control frame $f$ if the following conditions are both satisfied:
\begin{enumerate}
    \item the ball is outside the PZ of player A at frame $f+1$ and the ball displacement $\Delta {s_f}$ is above a given threshold, specified by the hyperparameter $\epsilon_s$,
    \item player A is not present on the subsequent control frame $f_c > f$ where there is either a possession or a duel.
\end{enumerate}
The first condition enables the algorithm to not detect as a loss situations where the ball remains static and the player moves without it, see Fig. \ref{fig:loss_gain}b$_1$. The second condition enables the detection of longer ball possessions by a player, where the ball eventually leaves the player's PZ but re-enters it after a number of frames where no other player has interacted with the ball, hence player A is still in possession and no loss is recorded, see Fig. \ref{fig:loss_gain}b$_2$. In all other circumstances, a possession loss is annotated, see Fig. \ref{fig:loss_gain}b$_3$-b$_4$.

\subsubsection{Possession gains}
Determining gains in possession requires asserting whether a given player not only is close to the ball, but also if they effectively make contact with it. Furthermore, since ball tracking data may lack $z$ information, additional logic is required to differentiate between actual possession gains and instances where the player(s) near the ball do not touch the ball. Following football intuition, we hypothesize that a change in both ball direction and ball speed is a strong indicator of players establishing contact with the ball, and thus gaining possession. 

For a given sequence of control frames $f_0,\ldots,f_n$ where the ball is within the PZ of the same player and $f_n\ge f_0$, we ascertain if there is an actual possession gain by introducing two hyperparameters, the minimum change in ball direction $\epsilon_{\theta}$ and minimum change in ball speed $\epsilon_v$. The ball is deemed to have changed direction within $[f_0,\,f_n]$ if the ball trajectory has changed from start to end of the control interval, namely ${\bf d}^0_{f_0}\cdot {\bf d}^1_{f_n}<\epsilon_\theta$; similarly, we consider the ball has changed speed if $\lbrace\abs{v^0_{f_i} - v^1_{f_i}}
>\epsilon_v\rbrace_{i=0}^n$ on at least one frame. All in all, we assume that if the ball has either changed direction or speed during a given control sequence $[f_0,\,f_n]$ involving the same player, then a possession gain occurs at $f_0$, see Fig. \ref{fig:loss_gain}c$_1$. Naturally, if both the ball trajectory and speed are not altered during a control frame interval, we consider that these control frames are false positives and not include them in the possession step, since it corresponds to instances where the ball travels near one or more players but none of them make explicit contact with the ball, see Fig. \ref{fig:loss_gain}c$_2$.

\subsubsection{Set piece triggers}

\begin{figure*}[h!]
    \centering
    \includegraphics[scale=.9]{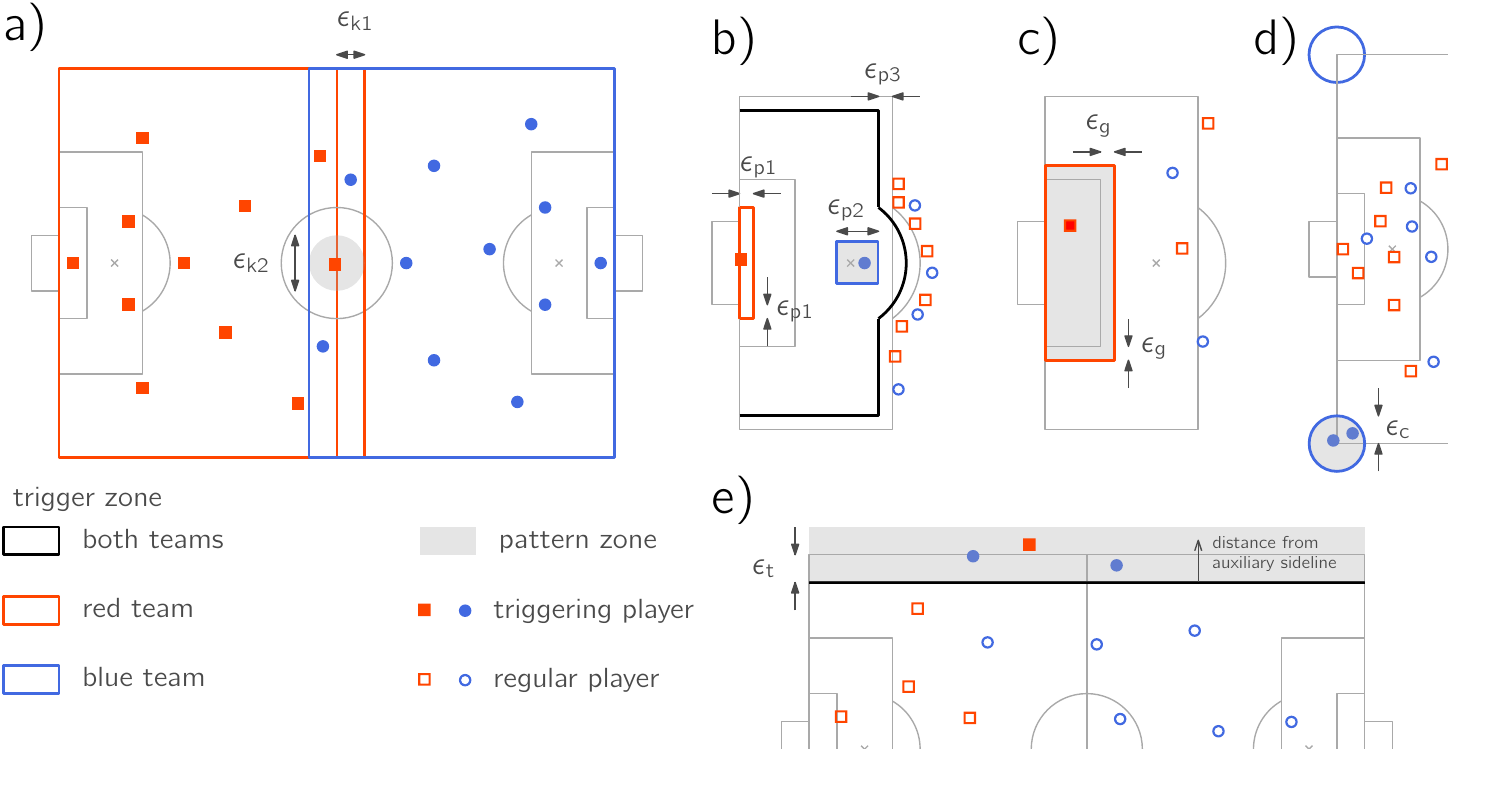}
    \caption{Schematic of set piece triggers (player configurations within the highlighted black/red/blue shape), triggering players (filled red/blue markers) and patterns (player in control of the ball within the grey shaded zones) for different set piece events: (a) Kickoff trigger with own half tolerance $\epsilon_{\rm k1}$; kickoff pattern with center mark tolerance $\epsilon_{\rm k2}$. (b) Penalty kick trigger with goal line tolerance $\epsilon_{\rm p1}$, no-player zone with tolerance $\epsilon_{\rm p3}$ and trigger and pattern with penalty mark tolerance $\epsilon_{\rm p2}$ and(c) Goal kick trigger and pattern with goal area tolerance $\epsilon_{\rm g}$. (d) Corner kick trigger and pattern with corner mark tolerance $\epsilon_{\rm c}$. (e) Throw-in trigger and pattern with sideline tolerance $\epsilon_{\rm t}$. Note that trigger and pattern zones coincide for corners, throw-ins and goal kicks.}
    \label{fig:db_trigger}
\end{figure*}

In addition to possession information, we incorporate several set piece triggers that inspect the spatial location of all players when the game is interrupted to determine which set piece event that resumes the game. For nomenclature purposes, we shall refer to the goal a team is attacking as \textit{active} goal, and analogously for other notable locations (penalty mark, corner mark, penalty area, goal area). The different triggers considered, along with tolerances to accommodate tracking data inaccuracies, are listed as follows:
\begin{itemize}
    \item dead ball trigger: the ball is dead, signaled by either a binary boolean or by the absence of ball tracking data.
    \item kickoff trigger: all players are within their own halves (with a tolerance of $\epsilon_{\rm k1}$) and there is at least one player within $\epsilon_{\rm k2}$ of the center mark, according to IFAB Law 8, see Fig. \ref{fig:db_trigger}a.
    \item penalty kick trigger: only one player is at their goal line between the posts (with tolerance bounding box of $\epsilon_{\rm p1}$), only one opponent is within a square bounding box from $\epsilon_{\rm p2}/4$ in front to $3\epsilon_{\rm p2}/4$ behind the active penalty mark, the other players are neither within the penalty area nor within \SI{9.15}{\m} from the penalty mark (with a tolerance of $\epsilon_{\rm p3}$), according to IFAB Law 14, see Fig. \ref{fig:db_trigger}b.
    \item goal kick trigger: at least one player is within their own goal area (with tolerance bounding box of $\epsilon_{\rm c}$), according to IFAB Law 16, see Fig. \ref{fig:db_trigger}c.
    \item corner kick trigger: at least one player is within $\epsilon_{\rm c}$ of one of their active corner marks, according to IFAB Law 17, see Fig. \ref{fig:db_trigger}d.
    \item throw-in trigger: at least one player is beyond the auxiliary sideline (sideline minus $\epsilon_{\rm t}$), according to IFAB Law 15, see Fig. \ref{fig:db_trigger}e.
\end{itemize}

The output of the possession step is the set of set piece triggers for each dead ball interval, since more than one may be triggered, as well as a table that features ball control (possession/duel) information and possession gains and losses. The former span multiple frames, whereas the latter are discretely annotated and will be mapped to football events in the event detection step described below.

\subsection{Event detection}\label{sec:events}
In this section, we discuss how the possession information and set piece triggers obtained in the previous one may be translated to both set piece and in-game football events.

\subsubsection{Set piece events}\label{sec:dbe}
The most straightforward segmentation of a football game is between in-game and dead ball intervals. A dead ball event (DBE) occurs immediately before a dead ball interval, and is followed by a set piece event (SPE) to resume the game. To that end, we establish the one-to-one correspondence between DBEs and SPEs detailed in Fig. \ref{fig:framework}b. Note that offsides are treated as fouls throughout this work, since from the tracking data perspective is a nontrivial task to distinguish between offsides and other infractions. Furthermore, we identify DBE-SPEs combining triggers and patterns, see Fig. \ref{fig:db_trigger}, as follows: (1) detect triggers in the spatial configuration of the players; (2) confirm DBE-SPE by ensuring the pattern is satisfied, namely the triggering player is within the pattern zone and the ball is within that player's possession zone on the first in-play frame. If there are multiple triggering players within the pattern zone and within $R_{pz}$ of the ball, we choose the closest player to the ball as the executor of the set piece event. Lastly, since free kicks lack distinct trigger configurations, we may only define a free kick pattern as a player having the ball within their possession zone on the first in-play frame. 

\begin{figure*}[h!]
    \centering
    \includegraphics[scale=1.5]{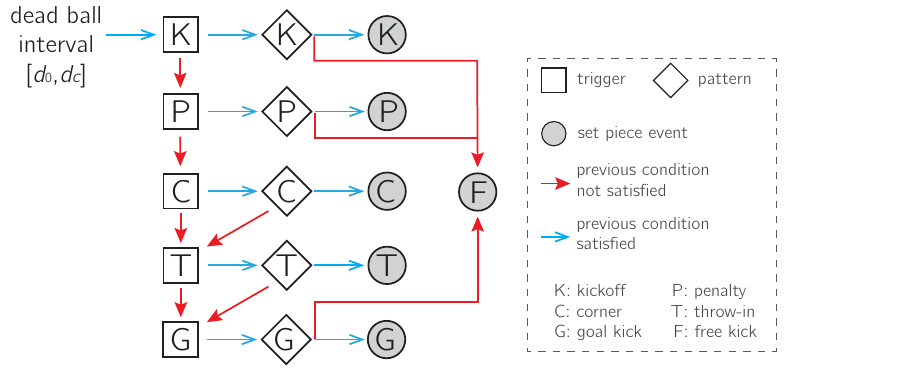}
    \caption{Flowchart to detect set piece events following a dead ball interval in the absence of tracking data errors.}
    \label{fig:db_flowchart}
\end{figure*}

For an arbitrary dead ball interval indexed by frames $[d_0,d_c]$, we examine if any of the set piece triggers are activated from an arbitrary intermediate frame $d_1$ ($d_0\le d_1\le d_c$) until the last dead ball frame $d_c$, hereafter referred to as complete triggers. The hierarchy established in Fig. \ref{fig:db_flowchart} is used to break ties between more than one complete triggers. Once a potential SPE using triggers has been identified, the algorithm aims to confirm it using the pattern. In the absence of tracking data inaccuracies, all set piece triggers that have been activated should be satisfied until the last dead ball frame $d_c$, and at the first in-play frame the ball should be at least within $R_{pz}$ of the set piece executor. If none of the patterns are satisfied, the algorithm will assume a free kick as a default option. The flowchart of this detection process is outlined in Fig. \ref{fig:db_flowchart}, where the set piece events are shown as grey circles and are always preceded in the auto-generated events table by the corresponding dead ball event. However, errors in tracking data impact the performance of this approach, and we refer the reader to Section S1 of Online Resource 1 for a detailed explanation on how to extend this framework if errors in tracking data are present.

Finally, two exceptions are accounted for regarding kickoffs. First, the one-to-one relation goal-kickoff no longer holds for last-minute goals whereby the period ends after the goal is scored and before the ball is kicked off. Therefore, a \texttt{goal?} dead ball event is added for shot sequences that cross the goal in the 2D plane at the end of the period, to express there is uncertainty whether a goal has been scored. Discerning whether these sequences are actual goals using only 2D tracking data is complex. In cases where the $z$-coordinate of the ball is available, the immediate solution would be to check whether the ball is below the crossbar when it crosses the goal line. 

Second, if during the game a kickoff is not properly executed the referee will order its repetition, which from the tracking data perspective could be mistaken as a distinct kickoff that would lead to an incorrect match score. To resolve this situation, assuming there are $k=1,\ldots,K$ kickoffs throughout a period, for each kickoff $k\ge2$ we check whether the ball has reached at least one of the penalty areas in the time interval between kickoff $k-1$ and kickoff $k$. If not, we assume the kickoff $k-1$ was mandated to be repeated and update the \textbf{from set piece} field to \texttt{incorrect kickoff} instead of \texttt{kickoff}, as well as the \textbf{dead ball event} occurring immediately prior to kickoff $k$ from \texttt{goal} to \texttt{referee interruption}.

In summary, the errors that we assume with this DBE detection process are due to the player-position tolerances we use for the triggers, as well as limitations of the tracking data: (1) throw-ins/free kicks occurring near the corner mark wrongfully classified as corner kicks; (2) free kicks occurring near the sidelines wrongfully classified as throw-ins; (3) free kicks occurring within the goal area wrongfully classified as goal kicks; (4) offsides being classified as fouls. Some of these errors could be circumvented by incorporating the $z$-coordinate of the ball for throw-ins, the pose of the referee or a video-based classifier.

\subsubsection{Shots and saves}\label{sec:shotsave}
The proposed framework of possession losses and gains extends naturally to the detection of shots and saves, arguably the most important in-game events. We define a shooting event as a possession loss by a player of the attacking team that is succeeded by a goal, a corner kick, a goal kick or a save. Furthermore, we define a saving event as a possession gain by the goalkeeper of the defending team where they are located inside the penalty area, and is preceded by a shooting event. Note that blocked shots are not encompassed in these definitions, since from the tracking data perspective a blocked shot is a possession loss succeeded by a possession gain of another player (either teammate or opponent) who is not the opposing goalkeeper. Blocked shots cannot therefore be associated with saving events, and hence will be identified as passes --either completed or intercepted. Another error that we assume are saves that occur after a ball deflection from a defender, since the algorithm will label them as a pass from the defender followed by a reception from the goalkeeper.

\begin{figure}[h!]
    \centering
    \includegraphics[scale=.8]{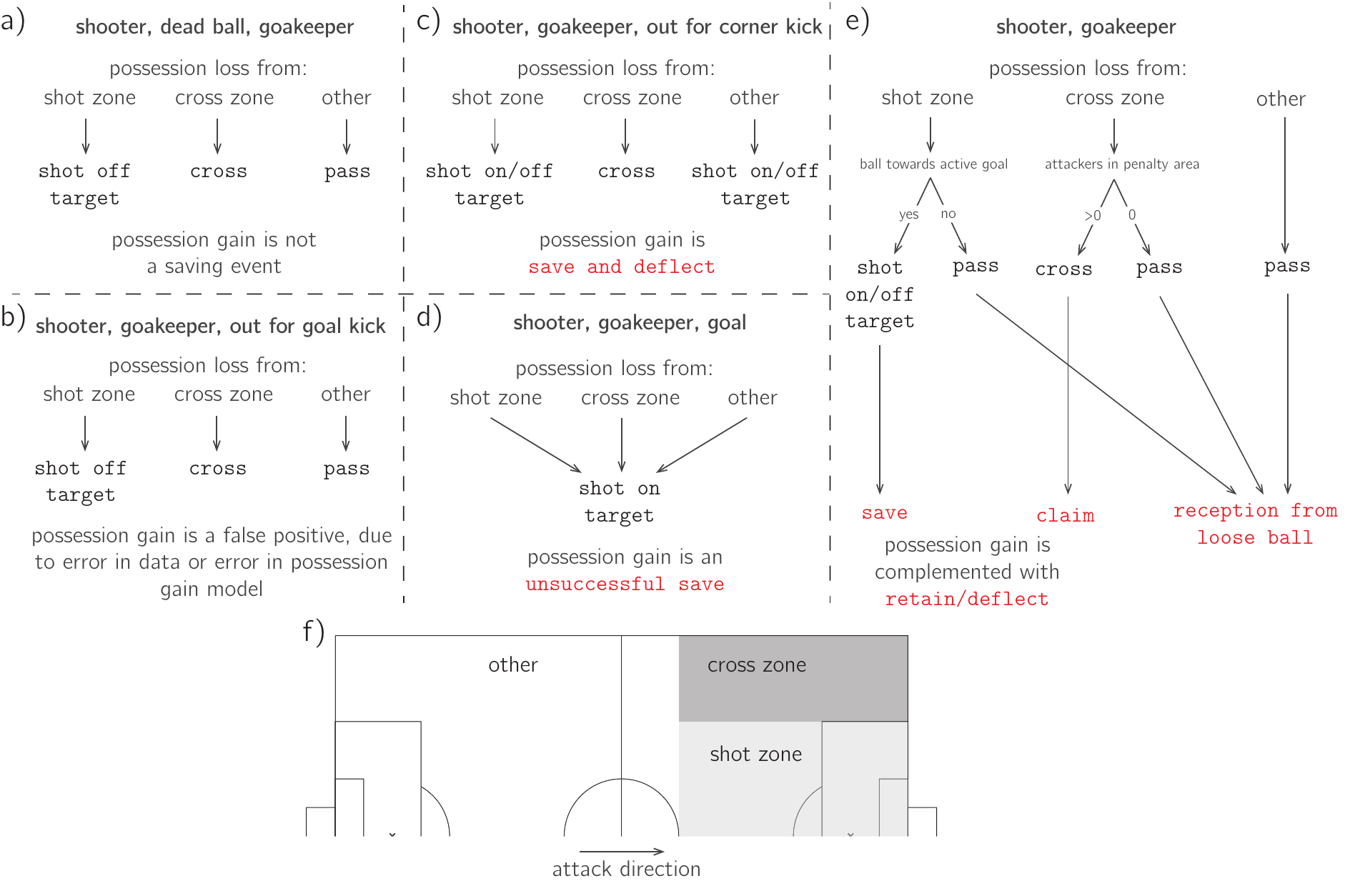}
    \caption{(a)-(e) Shot-save decision trees depending on the sequence of shooter, goalkeeper and dead ball. Shooting events are highlighted in black, whereas corresponding saving events are highlighted in red. (f) Sketch of football pitch distinguishing between cross zone and shot zone.}
    \label{fig:saveshot}
\end{figure}

For shooting events, we differentiate between shot on/off target, cross and pass. For saving events, we differentiate between save, claim, reception from a loose ball and unsuccessful save (a goal is conceded despite the goalkeeper touching the ball). The variables that are examined for each shot-save sequence are: whether a dead ball interval occurs before or after the goalkeeper's possession gain; the spatial location of the shooter, (crossing zone, shooting zone or other, see Fig. \ref{fig:saveshot}f); the direction of the ball after the possession loss occurs and whether it is moving towards the active goal; and the number of opponents in the penalty area. In addition, for the save/claim events we investigate if the goalkeeper loses possession within one second of the saving event, in order to distinguish between retention and deflection. Using these variables, we identify five main categories of shot-save sequences, which can be summarized in Fig. \ref{fig:saveshot} along with the distinct shooting (black) and saving (red) events that are extracted. The distinction between shot on/off target is made solely based on the ball trajectory immediately after the possession loss, with a \SI{0.25}{\m} tolerance beyond the goalposts.

\subsubsection{Crosses, receptions and interceptions}
Once the shots have been established, the distinction is made between crosses and passes using the same logic as described in Fig. \ref{fig:saveshot}. That is, for a possession loss to be labeled as a cross it needs to satisfy a few conditions: origin in the cross zone, the next player in control (attacking or defending) needs to be within the active penalty area and there should be at least one attacking player in the active penalty area. The remaining possession losses are therefore labeled as passes. Similarly, besides saving events the remaining possession gains are either receptions or interceptions, depending on whether the previous loss is made by a teammate or an opponent. 

Furthermore, football events can be complemented with several contextual attributes that can be computed or modeled from the tracking data, in the form of additional columns to the final events table. Examples include outcome of events, player location relative to the other team or number of opponents overtaken by passes \& possessions. We refer the reader to Section S5 of the Online Resource 1 for an exhaustive explanation on how contextual attributes are incorporated.

\section{\label{sec:results}Results}
In this section, we present the results of applying the above event detection framework to the datasets introduced in Section \ref{sec:data}. The chosen possession zone radii were $\SI{50}{\cm}$ for provider A and $\SI{1}{\m}$ for both provider B and provider C, whereas the radius of the duel zone were taken to be $R_{dz} = \SI{1}{\m}$ for all providers, see Section S3-S4 of Online Resource 1 for an extensive discussion on hyperparameters and tolerances.

In terms of computational resources, the datasets are stored in Google Cloud's BigQuery and we perform the computations on a Virtual Machine in Google Cloud featuring 1 CPUs and 4 GB of RAM. The code was developed in \texttt{Python 3}, and the mean computational wall time to execute the possession and event detection algorithm for a 90 minute match was three minutes.
%. and discuss potential applications that stem from automatic possession and event detection.

\subsection{Benchmarking with official event data}
\begin{figure}[h!]
    \centering
\includegraphics[scale=0.9]{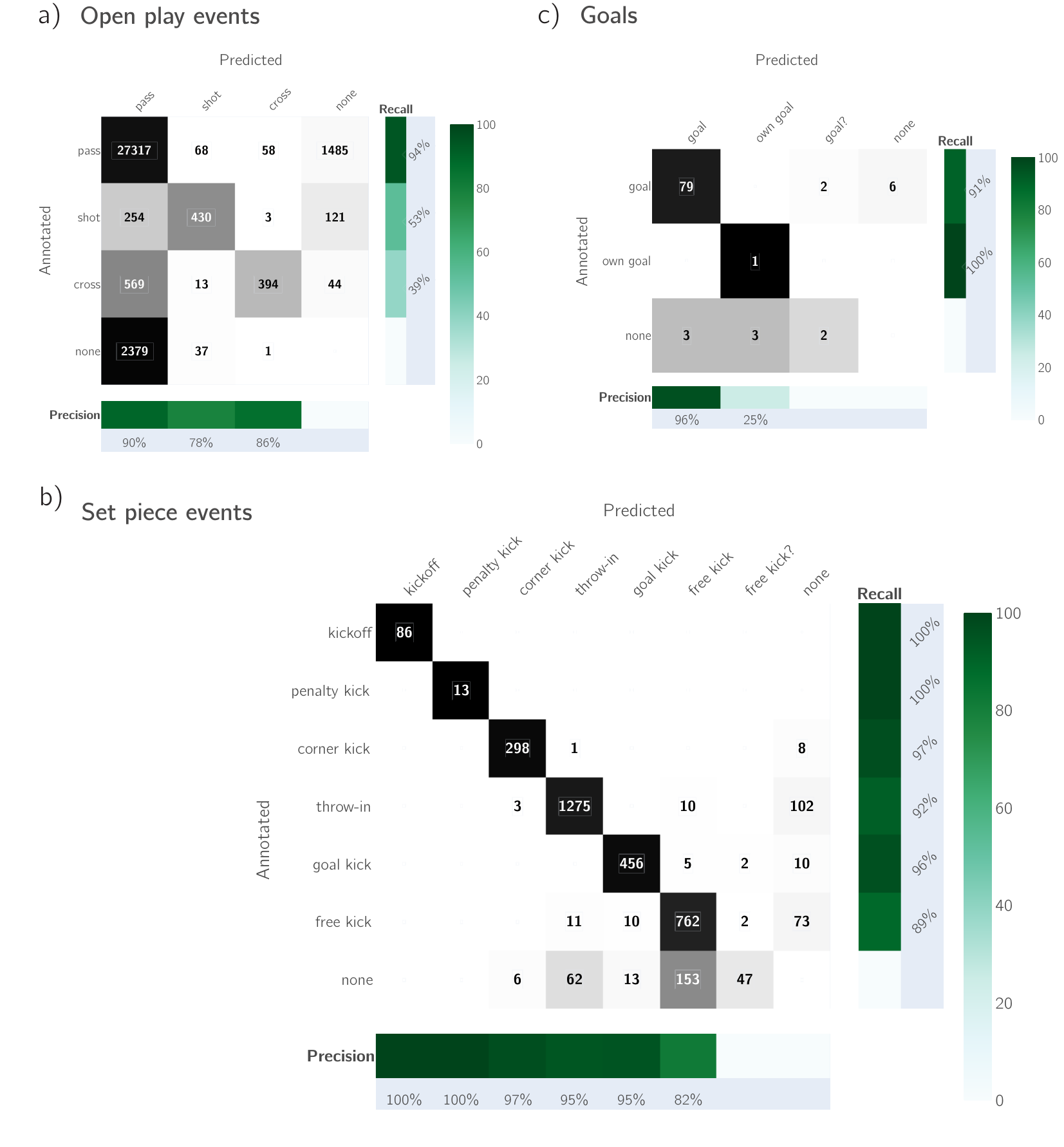}
    \caption{Confusion matrices comparing the events predicted by the event detection algorithm with the annotated events from STS, together with precision and recall for each category. Matrix cells are colored according to the relative number of instances per row. a) Open play passing events. b) Set piece events. c) Goals.}
    \label{fig:confusion}
\end{figure}

First, we present the results of benchmarking the automatically detected events with the manually annotated events by STS. The benchmarking criteria is outlined in Section S2 of Online Resource 1, and the results are shown with confusion matrices in Fig. \ref{fig:confusion}. We do not benchmark player possession data as this is not currently collected by event data providers. In addition, we should emphasize that manually collected event data is not without errors, hence detection rate can never be 100\%; we have observed several instances of non-annotated events, wrong timestamps or wrong players in the annotated events during the course of this research.

Open play passing events (passes, shots, crosses..) constitute the majority of events under consideration, with over 30K instances across all datasets. The confusion matrix, along with precision and recall for each category, is shown in Fig. \ref{fig:confusion}a. The most salient takeaway is the supra 90\% precision and recall in detecting passes. Furthermore, the shot precision was higher than the recall (78 vs 53) for a multitude of reasons: based on our definition of shot in Section \ref{sec:shotsave}, shots that were blocked by other players were labeled as passes by the algorithm (70\% of the 254 misclassifications), since the goalkeeper did not intervene; for shots that were not blocked, errors arise from situations with multiple players near the ball in which the shooter was wrongly identified. In addition, shots tended to take place on areas where players accumulate, hence it is not surprising that 15\% of shots (compared to less than 6\% of passes) were either not detected or attributed to another player. The main reason was that the tracking data (and consequently the event detection algorithm) exhibits more inaccuracies and errors when player occlusions occur. Regarding crosses, the main source of confusion were labeled crosses that the algorithm annotates as passes (569 misclassifications) based on the logic described in Fig. \ref{fig:saveshot}. Due to the absence of a gold standard definition of cross, these discrepancies are expected. 

The results for dead ball/set piece events are collected in Fig. \ref{fig:confusion}b. The ability to perfectly capture kickoffs and penalty kicks is paramount, since they constitute the best high-level descriptors of a game from the events perspective. The other categories for which a pattern exists (corner, throw-in, goal kick) also exhibit supra-90\% precision and recall, and the errors stem from mistakes in the inbounding player (e.g. more than one inbounding players are close, inbounding player is not tracked) and limitations of the tracking data (e.g. throw-in close to corner marks, free kick close to sideline or corner, free kick inside/near the goal area). Incorporating the ball $z$-coordinate would help in distinguishing throw-ins from corner kicks and free kicks. Finally, the worse results were for free kicks, which hold no specific pattern and were selected if no other spatial configuration was detected, as explained in Fig. \ref{fig:db_flowchart}. The presence of inaccuracies in player/ball tracking data discussed in Section S1 of Online Resource 1 lowers the precision of free kick detection as they were assigned to \texttt{free kick?}, which signals the algorithm was confused due to tracking data inaccuracies and requires external input.

The detection of goals is intrinsically related to the detection of kickoffs, whereby the goal (dead ball event) triggers a kickoff (set piece) as both the start and end of a dead ball interval. Nonetheless, benchmarking for goals separately allows us to analyze the performance of the proposed algorithm specifically on situations with many players involved (e.g. goal scored after a corner kick) where the algorithm may confuse a goal for an own goal, as well as goals scored at the end of the period (for which no kickoff pattern follows). The confusion matrix with the goal results is shown in Fig. \ref{fig:confusion}c, showing only six mistakes (goalscorer wrongly identified) and two last-minute goal events that did not correspond to a goal (algorithm unsure whether a goal was scored). Upon further inspection, the six goalscoring mistakes can be broken down as follows: the ball goes missing after the assist was made (3) and the data reflects an inaccurate situation (3). The two correctly matched last-minute goals correspond to the same late penalty kick goal, where tracking data from two different providers was available. The incorrectly detected last-minute goal corresponded to a shot that went above the crossbar, which could be corrected with the ball $z$-coordinate.

\subsection{Applications}\label{sec:applications}
In this section, we illustrate how both predicted event and possession information can be leveraged to perform statistical analyses. There is a plethora of different ways to slice and aggregate the event data, hence the choice largely depends on the objective of the study or the question that is put forth by the coach or analyst. A sample of potential analyses is presented below, which is by no means exhaustive and only intends to showcase how autodetected event and possession data may be used in a football analytics context. More applications can be found in Section S6 of Online Resource 1. For simplicity, the attacking direction is assumed to be from left to right. 

\subsubsection{Possession-informed player heatmap}
\begin{figure}[h!]
    \centering
    \includegraphics[scale=1]{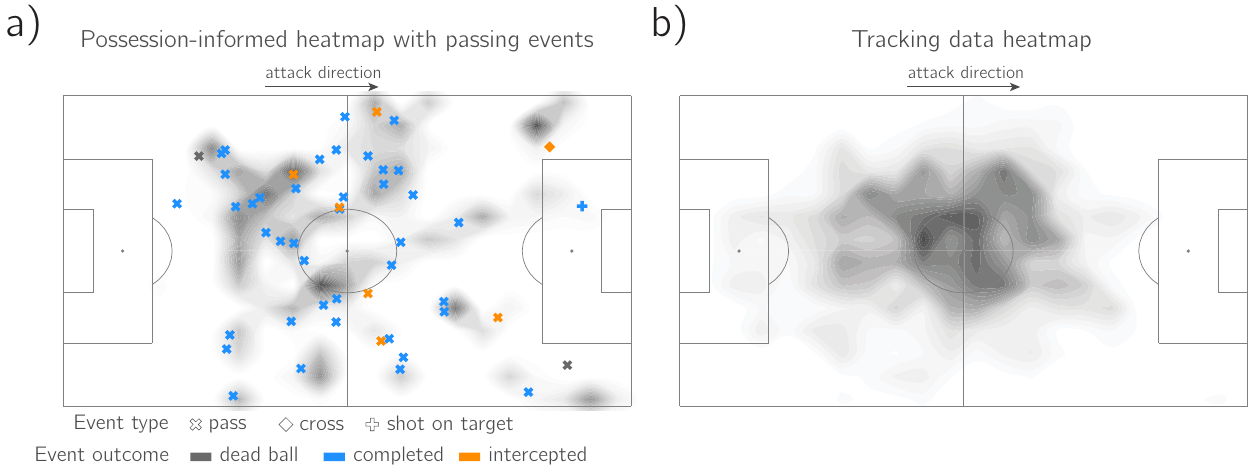}
    \caption{Heatmaps of spatial locations of a player during the first half of a game. a) Position of player only when the player is in possession, with passing events and outcomes overlaid. b) Position of player when ball is in-play, regardless of possession.}
    \label{fig:single_player}
\end{figure}

First, we choose one player on the first half of a game and visualize the heatmap of their locations when in possession, see Fig. \ref{fig:single_player}a, as well as the spatial distribution of passing events (distinguishing between passes, shots and crosses) and their outcome (completed, intercepted and dead ball), see Section S5 of Online Resource 1. The more traditional heatmap containing the player location at every in-play frame (where the player can be both with and without possession) is shown in Fig. \ref{fig:single_player}b for comparison. 

The main takeaway is that the possession-informed heatmap exhibits differences with respect to both the passing events distribution and the complete player heatmap, which signals the importance of capturing possession to more accurately understand the contribution of each player during the match. This approach can be seamlessy extended on many directions, e.g. composing the player heatmap when one of the teammates is in possession, when a specific opponent is in possession, or in a given interval of the match to name a few.

\subsubsection{Multiple match-aggregated event information}
We can aggregate and visualize data for the same team, player or both across multiple matches. The examples below correspond to a team for which we had data on five different games (two as the home team and three as the away team). We refer the reader to Section S5 and Fig. S4 of Online Resource 1 for further details on how the attributes discussed here (location of player, opponents overtaken, angle of passes, distance travelled by ball, pass origin) were evaluated.

\begin{figure}[h!]
    \centering
    \includegraphics[scale=1.1]{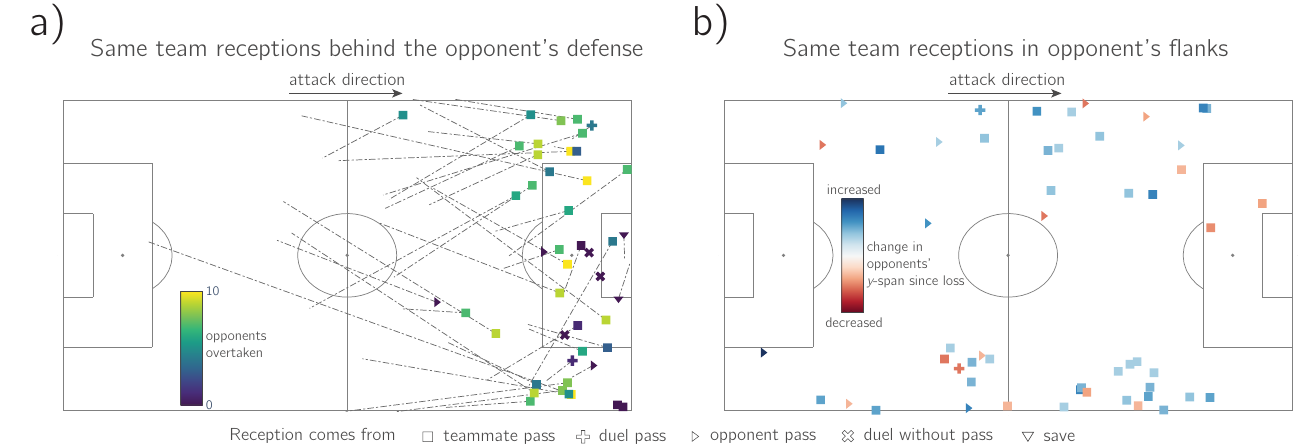}
    \caption{Scatter plot of receptions, where symbol refers to the nature of the prior passing event. a) Receptions behind the opponent's defense, colored by opponents overtaken by prior passing event. The ball trajectory from prior pass is shown in dash-dot. b) Receptions in the flanks of opponent, colored by change in $y$-span of opposing team between passing event and reception.}
    \label{fig:contextual}
\end{figure}

First, we analyze all receptions by a player on the team where the recipient is behind the opponents' defense --hence in a theoretically advantageous position to score-- along with depicting the pass trajectory, the nature of the event that lead to each reception and amount of opponents overtaken by it, see Fig. \ref{fig:contextual}a. Second, we examine the location of all receptions by a player on the flanks while illustrating the change in the opposing team's $x/y$-span between the prior passing event and the reception, see Fig. \ref{fig:contextual}b, where all flank receptions are colored by the change in y-span of the opposing team. 

\begin{figure}[h!]
    \centering
    \includegraphics[scale=0.85]{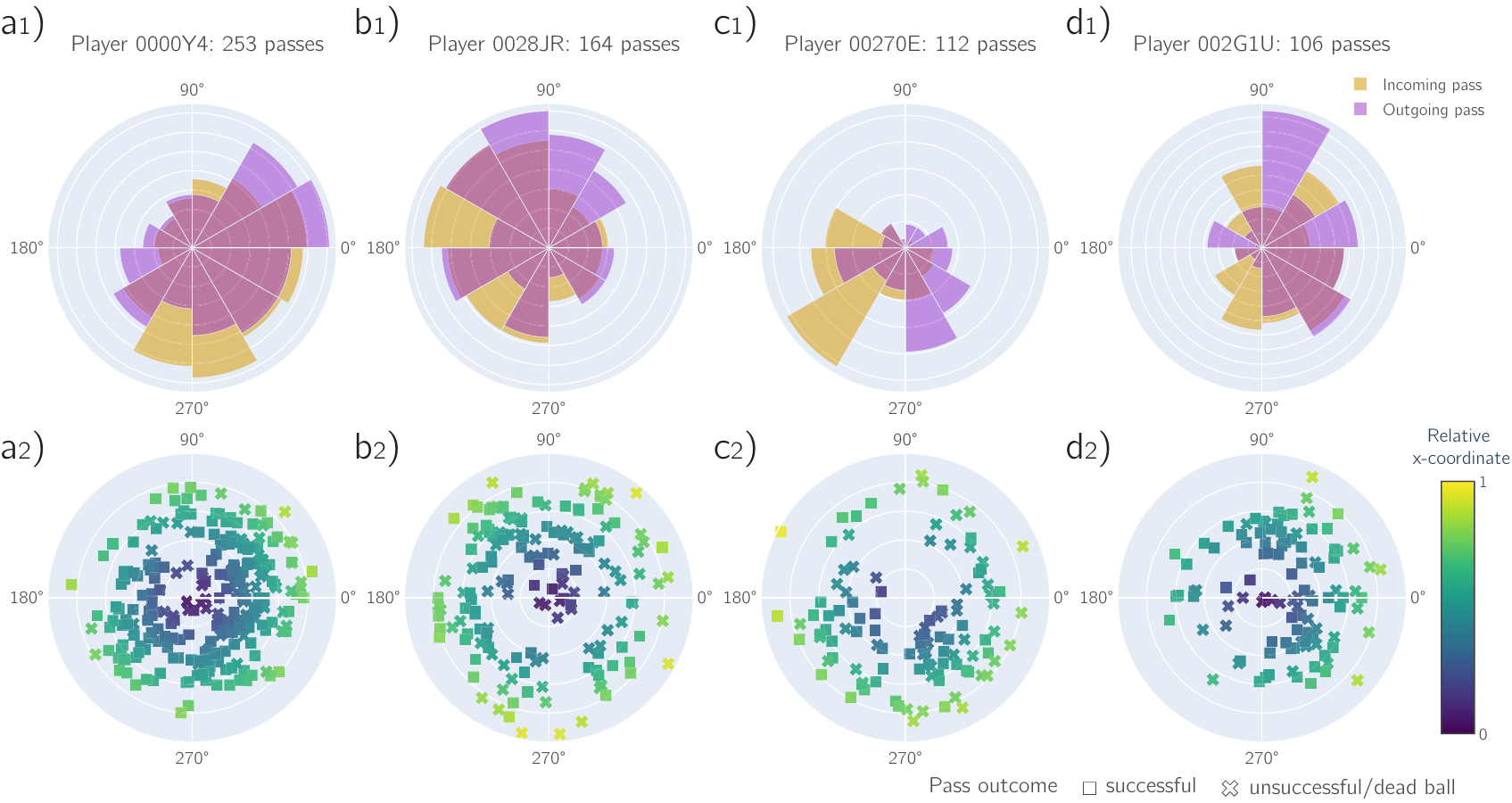}
    \caption{Angular information of ball trajectories for four midfielders with the most passes. a) Polar histogram of incoming and outgoing trajectories. b) Polar scatterplot of outgoing trajectories, where the symbol shows the outcome of the pass, the radial position shows how advanced was the player in the pitch when the pass was made (origin for own endline and outer circle for opposing endline) and the color refers to the distance traveled by the ball during the pass.}
    \label{fig:angles}
\end{figure}

Finally, we can investigate the trajectories of the passes by visualizing the incoming (at reception) and outgoing (at pass) trajectory angles of several players within the same team. The polar histogram of incoming and outgoing angles for the four midfielders with the most amount of passes is shown in the top row of Fig. \ref{fig:angles}a. Furthermore, polar scatterplots allow us to visualize all the outgoing angles for a given player (in the angular direction) while including information such as the outcome of the pass, how advanced was the player in the pitch when the pass was made (shown in the radial direction, circle origin for own endline and outer circle for opposing endline), and color-coded by the distance traveled by the ball from the time of the pass until reception/interception/out of bounds, see the bottom row of Fig. \ref{fig:angles}b. 

\section{Discussion}\label{sec:discussion}
In light of the these results, we can conclude that the proposed framework is effective in leveraging in-stadium tracking data to detect the majority (+90\%) of in-game and set piece events. However, as anticipated above the performance of the algorithm can be impacted by errors and availability of tracking data, errors in event data and modeling limitations, which are discussed below. 

The main limitation of the algorithm is that ball tracking data needs to be available, since we propose to detect events by assessing the change in distance between players and the ball. The other limitation is the absence of in-play/dead information, which is critical for set piece detection. Moreover, tracking data errors inevitably lead to wrongly predicted events, for instance player swaps or inaccurate in-play/dead ball boolean. Even though the available datasets feature accurate ball tracking data, namely ball-player distances at passing time are less than \SI{1}{\m} (see Section S3 of Online Resource 1), the proposed framework can be seamlessly applied to tracking data of lesser quality, for instance data collected from one tactical camera or from broadcast footage, by augmenting the possession zone radius and tuning the hyperparameters. Event data can also present several errors, for instance events not annotated, events attributed to a wrong player, or annotated event times more than \SI{10}{\s} before/after they occurred. These errors do not impact the auto-detected events, but they worsen the benchmarking results presented in Fig. \ref{fig:confusion}. 

From the modeling standpoint, the errors were due to the choice of parameters and hyperparameters or inherent limitations of the algorithm. For the former, we recommend a cross-validation strategy on a subset of matches to optimize the hyperparameter selection for each tracking data provider. For the latter, we identify several directions of improvement: (1) incorporating the $z$-coordinate of the ball; (2) the use of machine learning to identify events that are not rule-based, for instance blocked or deflected shots based on speed and context; (3) extending the possession zone definition to encompass a variable radius/shape based on pitch location, proximity of opponents and player velocity; (4) developing algorithms to extract pressure, team possession information as well as offensive and defensive configurations; (5) the incorporation of limb tracking data in addition to center-of-mass tracking data for all players and referees, with the objective of enhancing the granularity of already detected events (types of saves, body part for passes) while facilitating the detection of events that can be ambiguous from the tracking data perspective (tackles, types of duels, offsides, throw-in vs corner kick); (6) leveraging a synchronized audio feed that provides timestamped referee whistles to more accurately establish in-play/dead ball intervals; and (7) complementing the current approach with a video-based events classifier, which can enable the detection of refereeing events (cards, substitutions, VAR interventions) that are not captured by tracking data, in addition to improving the detection performance on edge-case set piece events, for instance drop-ball vs. free kick, corner kick vs. throw-in vs. free kick close to the corner marks; (8) applying the algorithm to broadcast tracking, which is less accurate than in-stadium tracking and the pitch is not always visible, which will thus require adjusting the algorithm's hyperparameters and dead ball patterns; (9) availability of additional datasets collected from different providers and stadiums to further test the validity of the proposed framework.

In terms of specific applications for the auto-generated event data, the broader context of the game encoded in the tracking data can be leveraged for a higher granular definition of the events. The examples introduced in Section \ref{sec:applications} and S6 of Online Resource 1 demonstrate how the generated possession and augmented event data may be used to perform advanced football analytics at the match, team and individual player level. We have introduced the notion of possession-informed heatmap to visually represent the locations of the player whilst only in possession of the ball, analyzed how our frame-to-frame ball possession information can be used to visualize possession distribution for both teams and among players, and finally showcased how the event data can be queried in search of highly specific events towards advanced analytics or video segmentation/selection, due to the auto-generated event data being in sync with the video and tracking data.

\section{Conclusions}
We have presented a decision tree-based computational framework that combines information on the spatial location of players and how the possession of the ball changes in time, both computed from 2D player and ball tracking data, with the laws of football to automatically generate possession and event data for a football match. The collection of event data is a manual, subjective and resource-intensive task, and is thus not available to most tournaments and divisions. The proposed framework is a suitable approach towards auto-eventing, due to the high accuracy (+90\%) observed, the limited computational burden and the ever-increasing availability and quality of tracking data feeds.

\section{Acknowledgements}
This research was conducted at the MIT Sports Lab and funded by FIFA through the MIT Pro Sports Consortium. The authors would like to acknowledge the founding partners of the MIT Sports Lab, Prof. Anette Hosoi and Christina Chase, for supporting this research effort. Automatic event detection with tracking data was initially explored as a class project in the context of the MIT Sports Lab class \textit{2.98/2.980 - Sports Technology: Engineering and Innovation}, by the team of students comprised of Juanita Becerra, Spencer Hylen, Steve Kidwell, Guillermo Larrucea, Kevin Lyons, \'{I}\~{n}igo de la Maza and Federico Ram\'{i}rez, under the supervision of Prof. Hosoi, Christina Chase and Ferran Vidal-Codina.  In addition, the authors would like to thank Eric Schmidt and Ramzi BenSaid from Google Cloud for the resources and help in leveraging the power of Google Cloud Platform. Finally, the authors thank Track160, DFL Deutsche Fu{\ss}ball Liga and Sportec Solutions for kindly providing part of the tracking and event data necessary to carry out this work.

\section{Conflict of interest statement}
One co-author serves as a Guest Editor for the Topical Collection for Football Research in Sports Engineering and another serves on the Editorial Board of Sports Engineering. Neither of them were involved in the blind peer review process of this paper. 

\section{Data availability statement}
The data used for this study was collected by FIFA at a number of its tournaments. Due to media and data rights, the datasets are not publicly available, but can be requested from \texttt{research@fifa.org} together with a viable research proposal.

% BibTeX users please use one of
%\bibliographystyle{sn-basic}      % mathematics and physical sciences
%\bibliographystyle{spmpsci}
\bibliographystyle{unsrtnat}

\bibliography{apssamp}% Produces the bibliography via BibTeX.

\end{document}